\documentclass[runningheads]{llncs}

\usepackage[final,year=2024,ID=10819]{eccv}

\usepackage{eccvabbrv}

\usepackage{graphicx}
\usepackage{booktabs}
\usepackage{multirow} 
\usepackage{makecell}
\usepackage{ulem}
\usepackage[linesnumbered,ruled,vlined]{algorithm2e}
\SetKwRepeat{Do}{do}{while}%
\usepackage[accsupp]{axessibility}  

\usepackage[pagebackref,breaklinks,colorlinks,citecolor=eccvblue]{hyperref}

\usepackage{orcidlink}

\begin{document}

\title{Domain Adaptation for Large-Vocabulary Object Detectors} 

\titlerunning{Abbreviated paper title}

\author{Kai Jiang \inst{1,\footnotemark[1]} \and
Jiaxing Huang \inst{2,\footnotemark[1]} \and
Weiying Xie\inst{1} \and
Jie Lei\inst{4} \and
Yunsong Li\inst{1} \and
Ling Shao\inst{3} \and
Shijian Lu\inst{2,\footnotemark[2]}
}

\authorrunning{F.~Author et al.}

\institute{State Key Laboratory of Integrated Services Networks, Xidian University, Xi’an 710071, China 
\email{kjiang\_19@stu.xidian.edu.cn}\\
\and
S-lab, School of Computer Science and Engineering, Nanyang Technological University
\email{\{Jiaxing.Huang, Shijian.Lu\}@ntu.edu.sg}\\
\and
UCAS-Terminus AI Lab, University of Chinese Academy of Sciences, Beijing, China \\
\and
School of Electrical and Data Engineering at the University of Technology Sydney
}

\footnotetext[1]{These authors contributed equally to this work.} 
\footnotetext[2]{Corresponding author.} 



\maketitle

\begin{abstract}
Large-vocabulary object detectors (LVDs) aim to detect objects of many categories, which learn super objectness features and can locate objects accurately while applied to various downstream data. However, LVDs often struggle in recognizing the located objects due to domain discrepancy in data distribution and object vocabulary. At the other end, recent vision-language foundation models such as CLIP demonstrate superior open-vocabulary recognition capability. 
This paper presents KGD, a Knowledge Graph Distillation technique that exploits the implicit knowledge graphs (KG) in CLIP for effectively adapting LVDs to various downstream domains.
KGD consists of two consecutive stages: 1) KG extraction that employs CLIP to encode downstream domain data as nodes and their feature distances as edges, constructing KG that inherits the rich semantic relations in CLIP explicitly; 
and 2) KG encapsulation that transfers the extracted KG into LVDs to enable accurate cross-domain object classification. 
In addition, KGD can extract both visual and textual KG independently, providing complementary vision and language knowledge for object localization and object classification in detection tasks over various downstream domains. 
Experiments over multiple widely adopted detection benchmarks show that KGD outperforms the state-of-the-art consistently by large margins.
  \keywords{Domain adaptation \and Large-vocabulary object detectors \and Vision-language models \and Knowledge graph distillation}
\end{abstract}

\section{Introduction}
\label{sec:intro}



Object detection aims to locate and classify objects in images, which conveys critical information about ``what and where objects are'' in scenes. It is very important in various visual perception tasks in autonomous driving, visual surveillance, object tracking, etc. Unlike traditional object detection, 
large-vocabulary object detection~\cite{li2022grounded,yao2022detclip, zhou2022detecting} 
aims to detect objects of a much larger number of categories, e.g., 20k object categories in~\cite{zhou2022detecting}. It has achieved very impressive progress recently thanks to the availability of large-scale training data. 
On the other hand, large-vocabulary object detectors (LVDs) often struggle while applied to various downstream tasks as their training data often have different distributions and vocabularies as compared with the downstream data, i.e., due to domain discrepancies.

\begin{figure}[tb]
\centering
\includegraphics[width=0.85\linewidth]{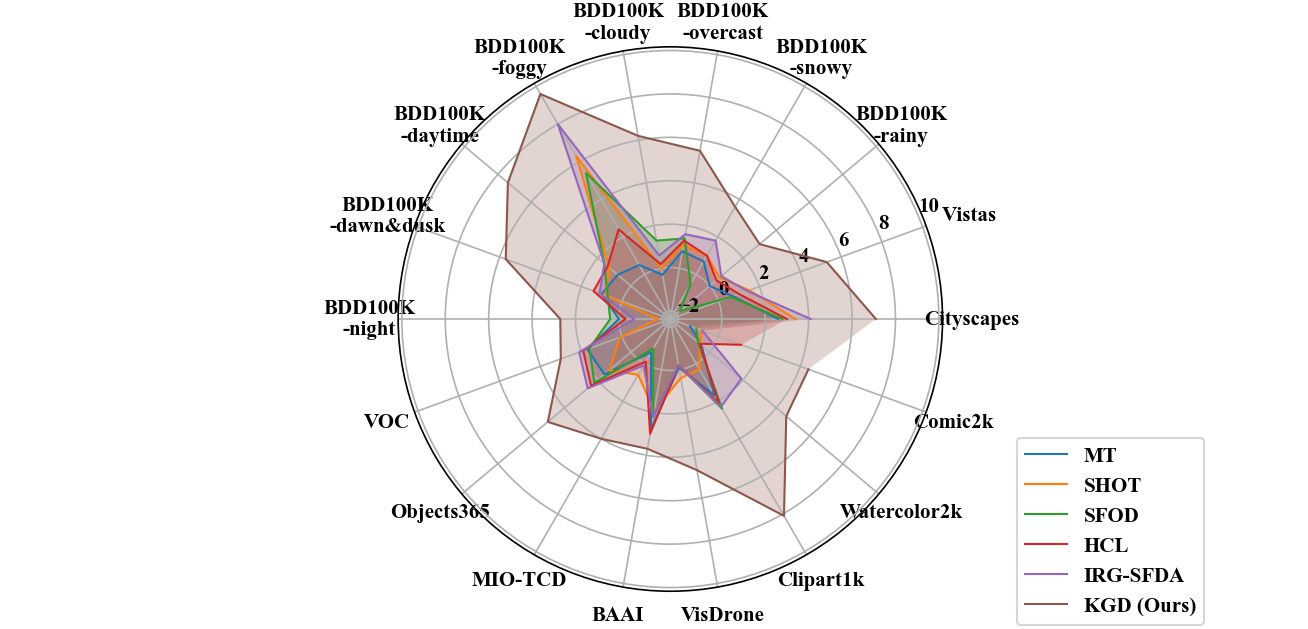}
\caption{
A comparison of the domain adaptation performance of our method against existing methods. 
Our method outperforms the state-of-the-art consistently on 11 widely studied downstream detection datasets in terms of AP50 improvements.
The results of all methods are acquired with the same baseline~\cite{zhou2022detecting}.
}
\label{RadarGraph}
\vspace{-2.0em}
\end{figure}

In this work, we study unsupervised domain adaptation of LVDs, i.e., how to adapt LVDs towards various downstream tasks with abundant unlabelled data available. Specifically, we observe that LVDs learn superb generalizable objectness knowledge from massive object boxes, being able to locate objects in various downstream images accurately~\cite{zhou2022detecting}. However, LVDs often fail to classify the located object due to two major factors: 
1) the classic dataset-specific class-imbalance and the resultant distribution bias across domains; 
and 2) different vocabularies across domains~\cite{oksuz2020imbalance, you2019universal}. 
At the other end, vision-language models (VLMs)~\cite{zhang2023visionlanguage} such as CLIP~\cite{radford2021learning} learn from web-scale images and text of arbitrary categories, which achieve significant generalization performance 
in various downstream tasks with severe domain shifts. 
Hence, effective adaptation of LVDs towards various unlabelled downstream domains could be facilitated by combining the superior object localization capability from LVDs and the super-rich object classification knowledge from CLIP.

We design Knowledge Graph Distillation (KGD) that explicitly retrieves the classification knowledge of CLIP to adapt LVDs while handling various unlabelled downstream domains. 
KGD works with one underling hypothesis, i.e., the generalizable classification ability of CLIP largely comes from its comprehensive knowledge graph learnt over billions of image-text pairs, which enables it to classify objects of various categories accurately. In addition, the knowledge graph in CLIP is implicitly encoded in its learnt parameters which can be exploited in two steps: 1) Knowledge Graph Extraction (KGExtract) that employs CLIP to encode downstream data as nodes and computes their feature distances as edges, constructing an explicit CLIP knowledge graph that captures inherent semantic relations as learnt from web-scale image-text pairs; and 2) Knowledge Graph Encapsulation (KGEncap) that encapsulates the extracted knowledge graph into object detectors to enable accurate object classification by leveraging relevant nodes in the CLIP knowledge graph.

The proposed KGD allow multi-modal knowledge distillation including Language Knowledge Graph Distillation (KGD-L) and Vision Knowledge Graph Distillation (KDG-V). Specifically, KGD-L considers texts as nodes and the distances among text embeddings as edges, enabling detectors to reason whether a visual object matches a text by leveraging other relevant text nodes. KGD-V takes a category of images as a node and the distances among image embeddings as edges, which enhances detection by conditioning on other related visual nodes. Hence, KGD-L and KGD-V complement each other by providing orthogonal knowledge from language and vision perspectives.
In this way, KGD allows to explicitly distill generalizable knowledge from CLIP to facilitate unsupervised adaptation of large-vocabulary object detectors towards distinctive downstream datasets.

In summary, the major contributions of this work are threefold. \textit{First}, we propose a knowledge transfer framework that exploits CLIP for effective adaptation of large-vocabulary object detectors towards various unlabelled downstream data. To the best of our knowledge, this is the first work that studies distilling CLIP knowledge graphs for the object detection task. \textit{Second}, we design novel knowledge graph distillation techniques that extracts visual and textual knowledge graphs from CLIP and encapsulates them into object detection networks successfully. \textit{Third}, extensive experiments show that KGD outperforms the state-of-the-art consistently across 11 widely studied detection datasets as shown in Fig.~\ref{RadarGraph}.

\section{Related works}

\textbf{Large-vocabulary Object Detection}~\cite{dave2021evaluating,gupta2019lvis,redmon2017yolo9000, yang2019detecting} 
aims to detect objects of thousands of classes. 
Most existing studies tackle this challenge by designing various class-balanced loss functions~\cite{dave2021evaluating} for effective learning from large-vocabulary training data and handling the long-tail distribution problem~\cite{li2020overcoming,feng2021exploring,wu2020forest,zhang2021distribution}.
Specifically, several losses have been proposed, such as Equalization losses~\cite{tan2020equalization, tan2021equalization}, SeeSaw loss~\cite{wang2021seesaw}, and Federated loss~\cite{zhou2021probabilistic}.  
On the other hand, \cite{Yang_2019_ICCV} and Detic~\cite{zhou2022detecting} attempt to introduce additional image-level datasets with large-scale fine-grained classes for training large-vocabulary object detector (LVD), aiming to expand the detector vocabulary to tens of thousands of categories. 
These LVDs learn superb generalizable objectness knowledge from object boxes of massive categories and are able to locate objects in various downstream images accurately~\cite{zhou2022detecting}. 
However, they often fail to classify the located objects~\cite{oksuz2020imbalance, you2019universal} accurately.
In this work, we focus on adapting LVDs towards various unlabelled downstream data by utilizing the super-rich object classification knowledge from CLIP.

{
\textbf{Domain Adaptation} aims to adapt source-trained models towards various target domains.
Previous work largely focuses on unsupervised domain adaptation (UDA), which minimizes the domain discrepancy by discrepancy minimization~\cite{long2015learning, vu2019advent}, adversarial training~\cite{gong2019dlow, vu2019advent, luo2021category, sun2023domain,munir2021ssal}, self-supervised learning~\cite{chen2023pipa,yue2021prototypical,wang2021domain}, or self-training~\cite{lee2013pseudo, zhang2019category, zhang2021prototypical,mei2020instance,zou2021geometry,liu2021cycle,chen2020self,munir2021ssal,yang2021st3d,kim2019self,zou2018unsupervised}.
Recently, source-free domain adaptation (SFDA) generates pseudo labels for target data without accessing source data, which performs domain adaptation with entropy minimization~\cite{liang2020we}, self training~\cite{tarvainen2017mean, li2021free,karim2023c,yi2023source,qu2022bmd}, contrastive learning~\cite{huang2021model, vs2022instance,lo2023spatio,zhang2022divide}, etc. 
However, most existing domain adaptation methods struggle while adapting LVDs toward downstream domains, largely due to the low-quality pseudo labels resulting from the discrepancy in both data distributions and object vocabulary.
}

\textbf{Vision-Language Models (VLMs)} have achieved great success in various vision tasks~\cite{zhang2023visionlanguage}. They are usually pretrained on web-crawled text-image pairs with a contrastive learning objective. Representative methods such as CLIP~\cite{radford2021learning} and ALIGN~\cite{jia2021scaling} have demonstrated very impressive generalization performance in many downstream vision tasks. Following~\cite{radford2021learning,jia2021scaling}, several studies~\cite{jia2021scaling, kim2021vilt, yao2021filip, li2022blip} incorporate cross-attention layers and self-supervised objectives for better cross-modality modelling of noisy data. In addition, several studies~\cite{fürst2022cloob, doveh2022teaching, pei2022hierarchical, gao2022pyramidclip} learn fine-grained and structural alignment and relations between image and text. 
In this work, we aim to leverage the generalizable knowledge learnt by VLMs to help adapt LVDs while handling various unlabelled downstream data.

\textbf{Knowledge Graph (KG)}~\cite{peng2023knowledge} is a semantic network that considers real-world entities or concepts as nodes and treats the semantic relations among them as edges.
Multi-modal knowledge graph~\cite{alberts2020visualsem, zhu2022multi} extends knowledge from text to the visual domain, enhancing machines' ability to describe and comprehend the real world. 
These KGs have proven great effectiveness in storing and representing factual knowledge, leading to successful applications in various fields such as entity recognition~\cite{zhang2018adaptive, wilcke2020end}, question-answering~\cite{marino2021krisp}, 
and information retrieval~\cite{deng2021gakg}. 
Different from the aforementioned KGs and MMKGs that are often handcrafted by domain experts, we design knowledge graph distillation that builds a LKG and a VKG by explicitly retrieving VLM's generalizable knowledge learnt from web-scale image-text pairs, which effectively uncover the semantic relations across various textual and visual concepts in different downstream tasks, ultimately benefiting the adaptation of LVDs.

\begin{figure}[t]
\centering
\includegraphics[width=1.0\linewidth]{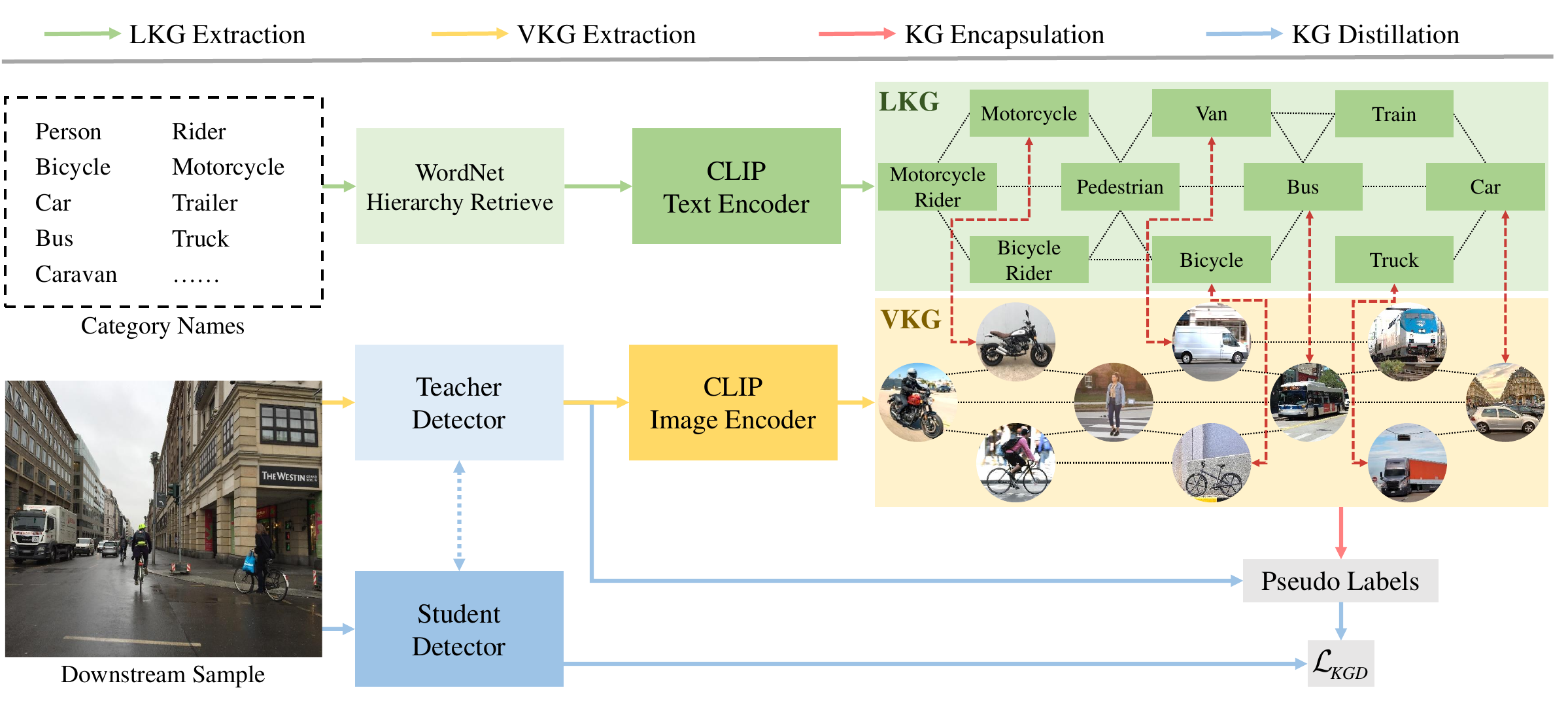}
\caption{
Overview of the proposed Knowledge Graph Distillation (KGD). 
KGD comprises two consecutive stages including Knowledge Graph Extraction (KGExtract) and Knowledge Graph Encapsulation (KGEncap).  
KGExtract employs CLIP to encode downstream data as nodes and considers their feature distances as edges, explicitly constructing KGs including language knowledge graph (LKG) and vision knowledge graph (VKG) that inherit the rich semantic relations in CLIP. 
KGEncap transfers the extracted KGs into the large-vocabulary object detector to enable accurate object classification over downstream data. 
Besides, KGD works for both image and text data and allow extracting and transferring vision KG (VKG) and language KG (LKG), providing complementary knowledge for adapting large-vocabulary object detectors for handling various unlabelled downstream domains.
}
\label{Schematic}
\vspace{-2.0em}
\end{figure}

\section{Method}

\textbf{Task Definition.} This paper focuses on unsupervised adaptation of large-vocabulary object detectors (LVDs).
We are provided with a set of unlabeled downstream domain data $\mathcal{D}_t= \{\mathbf{x}_i^t\}_{i=1}^{N_t}$ and an LVD pre-trained on labeled source domain detection dataset $\mathcal{D}_s=\{\mathbf{x}_i^s, \mathbf{y}_{i}^{s}\}_{i=1}^{N_s}$. 
$\mathbf{x}_i$ and $\mathbf{y}_i=\{(\mathbf{p}_j,\mathbf{t}_j)\}_{j=1}^M$ are the image and $M$ instance annotations of $i$-th sample, where $\mathbf{p}_j$ and $\mathbf{t}_j$ denote the ground-truth category and box coordinate of $j$-th instance.
$N_s$ and $N_t$ refer to the number of samples in $\mathcal{D}_s$ and $\mathcal{D}_t$. 
The goal is to adapt the pretrained LVD towards the downstream domain $\mathcal{D}_t$ by using the unlabelled images.

\textbf{Na\"ive Solution with Mean Teacher Method (MT)~\cite{tarvainen2017mean}.}
In this paper, we adopt Detic~\cite{zhou2022detecting} as the pretrained LVD, which utilizes CLIP text embeddings as the classifier.
We employ mean teacher~\cite{tarvainen2017mean} as the preliminary solution, which involves a teacher detector and a student detector where the former generates pseudo labels to train the latter while the latter updates the former in a momentum manner.
Given a batch of $B$ unlabeled target samples, the teacher detector $\Phi_t$ first produces detection predictions on them, which are then filtered with a predefined threshold $\tau$ to generate detection pseudo label $\hat{\mathbf{y}}_i$ (consisting of classes and bounding boxes). With $\hat{\mathbf{y}}_i$, the unsupervised training of student detector $\Phi_s$ on the unlabeled downstream data can be formulated as the following:
\begin{equation}\label{baselinefun}
    Loss = \frac{1}{B} \sum_{i=1}^{B}\mathcal{L}\left(\Phi_s(\mathbf{x}_i^t), \hat{\mathbf{y}}_i\right),
\end{equation} 
where  
$\mathcal{L}(\cdot)=\mathcal{L}_{rpn}(\cdot)+\mathcal{L}_{reg}(\cdot)+\mathcal{L}_{cls}(\cdot)$ is the detection loss function in which $\mathcal{L}_{rpn}(\cdot)$, $\mathcal{L}_{reg}(\cdot)$, and $\mathcal{L}_{cls}(\cdot)$ denote the loss for region proposal network, regression, and classification, respectively. Note both teacher detector $\Phi_t$ and student detector $\Phi_s$ are initialized with the pretrained LVD.

\textbf{Motivation.} On the other hand, although the LVD is able to locate objects in various downstream-domain images accurately~\cite{zhou2022detecting}, it often fails to classify the located objects, leading to very noisy detection pseudo labels when serving as the teacher detector.
At the other end, vision-language models (VLMs)~\cite{zhang2023visionlanguage} such as CLIP~\cite{radford2021learning} learns from web-scale images-text pairs of arbitrary categories, which possesses the ability to classify objects accurately in various downstream data.
Thus, we argue that effective adaptation of LVDs towards various unlabelled downstream data could be facilitated by combining the superior object localization capability from LVDs and the super-rich object classification knowledge from CLIP.
To this end, we design Knowledge Graph Distillation (KGD) with Language KGD and Vision KGD, aiming to explicitly retrieves the classification knowledge of CLIP to adapt LVDs while handling various unlabelled downstream data.
The overview of our proposed KGD is shown in Fig.~\ref{Schematic}.

\subsection{Language knowledge graph distillation}
\label{KGD-L}

The proposed language knowledge graph distillation (KGD-L) aims on distilling knowledge graph from the perspective of text modality.
KGD-L works in a two-step manner. 
The first step is language knowledge graph (LKG) extraction with a large lexical database named WordNet~\cite{miller1995wordnet} that aims to uncover the implicitly encoded language knowledge in CLIP.
With the guidance from the WordNet that stores a wide range of knowledge, LKG Extraction builds a category-discriminative and domain-generalizable LKG.
The second step is LKG encapsulation that encapsulates the extracted LKG into the teacher detector, enabling the detector to reason whether a visual object matches a text by leveraging other relevant text nodes and ultimately generate more accurate detection pseudo labels.

\textbf{LKG Extraction with WordNet Hierarchy.}
We first generate domain-generalizable prompts for each object category by leveraging the large lexical database WordNet~\cite{miller1995wordnet}.
Specifically, given the category set $\mathcal{C}=\{\mathbf{c}_i|i=1…,N_c\}$ of 
a downstream domain, 
we obtain the WordNet~\cite{miller1995wordnet} Synset definition as well as the hyponym set of category $\mathbf{c}_i$ as follows:
\begin{equation}\label{WordNetDefGen}
    \mathbf{d}_i,\mathcal{S}_i=\text{WNRetrieve}(\mathbf{c}_i),
\end{equation}
where $\text{WNRetrieve}(\cdot)$ retrieves the WordNet database~\cite{miller1995wordnet} and returns the definition $\mathbf{d}_i$ as well as the hyponym set $\mathcal{S}_i$ of its input. 
$\mathcal{S}_i=\{\mathbf{s}_j\}_{j=1}^m$, where $\mathbf{s}_j$ refers to the $j$th hyponym of category $\mathbf{c}_i$ and $m$ refers to the cardinal number of $\mathcal{S}_i$.
In this way, a category name $\mathbf{c}_i$ can be better defined and described with the informative yet accurate category definition in its hyponym set from WordNet, 
which are then combined with $\mathbf{d}_i$ as a set of domain generalizable prompts for category $\mathbf{c}_i$:
\begin{equation}\label{MultipleDescriptionCateSet}
    \tilde{\mathcal{S}_i}=\mathcal{S}_i \cup \{\mathbf{d}_i\},
\end{equation}
and the domain generalizable prompt set of category set $\mathcal{C}$ can be constructed as the following:
\begin{equation}\label{MultipleDescriptionSet}
    \tilde{\mathcal{S}}=\mathop{\cup}_{i=1}^{N_c}\tilde{\mathcal{S}_i}.
\end{equation}
With the category-discriminative and domain-generalizable information contained in $\tilde{\mathcal{S}}$, we formulate the proposed LKG as a weighted undirected graph $G_{L} = ({V}_L, {U}_L, \Omega)$
, which is capable of capturing semantic relationships and associations between different category concepts.
${V}_L =\{\tilde{\mathbf{s}}_i\}_{i=1}^{N_c(m+1)}$ is the vertex set in which each node $\tilde{\mathbf{s}}_i$ refers to a description in $\tilde{\mathcal{S}}$.
And ${U}_L =\{(\tilde{\mathbf{s}}_i,\tilde{\mathbf{s}}_j)\}$ is the edge set.
$\Omega$ is a matrix of node feature vectors $\Omega_i=T(\tilde{\mathbf{s}}_i)$, where $T(\cdot)$ denotes the CLIP text encoder.

\textbf{LKG Encapsulation} encapsulates the comprehensive knowledge in the extracted LKG into the teacher detector to facilitate detection pseudo label generation.
Specifically, we first employ CLIP to encode the regions cropped by the teacher detector and then generate pseudo labels for each region feature conditioned on LKG.
Given the image $\mathbf{x}^t \in \mathcal{D}_t$, we feed it into the teacher detector $\Phi_t$ to acquire the prediction as the following:
\begin{equation}\label{PseudoLabelGen}
    \hat{\mathbf{y}}={\Phi}_t(\mathbf{x}^t),
\end{equation}
where $\hat{\mathbf{y}}=\{(\hat{\mathbf{p}}_j,\hat{\mathbf{t}}_j)\}_{j=1}^M$,
$\hat{\mathbf{p}}_j $ denotes the probability vector of the predicted bounding box $\hat{\mathbf{t}}_j$ after Softmax activation function. 
$M$ denotes the number of predicted proposals after the thresholding with $\tau$, i.e., a predicted proposal will be discarded if its confidence score is less than $\tau$.

Next, we employ CLIP to encode the predicted object proposals in $\hat{\mathbf{y}}$ as follows:
\begin{equation}\label{VLMFeatureGen}
    {F}=V\left(Crop\left(\mathbf{x}^t,\hat{\mathbf{y}}\right)\right),
\end{equation}
where $Crop(\cdot)$ crops square regions from image $\mathbf{x}^t$ based on the longer edges of bounding boxes in $\hat{\mathbf{y}}$, $V(\cdot)$ is the image encoder of CLIP, and the $j$-th 
vector $\mathbf{f}_j$ of matrix ${F}$ is the feature of $j$-th proposal in $\hat{\mathbf{y}}$.

With the extracted LKG $G_{L}$ and the features of objects (or object proposals) $F$, 
we reason the class of objects conditioned on $G_{L}$ with a two-layer graph convolutional network (GCN)~\cite{wu2020comprehensive} as follows:
\begin{equation}
    [Q^F; Q^{\Omega}] = \text{Softmax}(D^{-\frac{1}{2}} {A} D^{-\frac{1}{2}}\;\text{ReLU}(D^{-\frac{1}{2}} {A} D^{-\frac{1}{2}} H^0 W^0)W^1),
\end{equation}
where 
$H^0=[F; \Omega]$, 
$A_{ij}=exp(-||H^0_{i}-H^0_{j}||_2^2/\text{Var}(||H^0_{i}-H^0_{j}||_2^2))$, $A_{ii}=1$, 
and $D_{ii}=\sum_jA_{ij}$.
$Q^F_{ji}/Q^{\Omega}_{ji}$ is the $i$-th element in probability vector $Q^F_{j}/Q^{\Omega}_{j}$, which denotes the predicted category probability of being $\mathbf{c}_i$ for object feature $\mathbf{f}_j$/LKG node $\tilde{\mathbf{s}}_j$.
$\{W^l\}_{l=0}^{1}$ are the trainable weights.
For updating
$\{W^l\}_{l=0}^{1}$, 
we minimizing the following cross entropy error over the nodes in LKG:
\begin{equation}
    \mathcal{L}_{LKG}(\mathbf{x}^t) = -\sum_{i}\sum_{j} \left(log(Q^{\Omega}_{ji}) \cdot \mathbb{I}(\tilde{\mathbf{s}}_j\in \tilde{S}_i)\right).
\end{equation}

Then we encapsulate the extracted LKG into $\Phi_t$ by,
\begin{equation}
\label{L-KGMeasurement}
    {\mathbf{p}}_{ji}^{l} = \hat{\mathbf{p}}_{ji} \cdot Q^F_{ij},
\end{equation}
where $\hat{\mathbf{p}}_{ji}$ is the $i$-th element in probability vector $\hat{\mathbf{p}}_j$, which denotes the predicted category probability of $\mathbf{c}_i$.
The first term in Eq.~\ref{L-KGMeasurement} denotes the original prediction probability from the teacher model while the second term in Eq.~\ref{L-KGMeasurement} stands for the prediction probability from LKG. ${\mathbf{p}}_{ji}^{l}$ denotes the prediction probability calibrated by LKG.

In this way, KGD-L extracts and encapsulates LKG from CLIP into the teacher detector, enabling it to reason whether an object matches a category conditioned on the relevant nodes in LKG and ultimately refining the original detection pseudo labels.

\subsection{Vision knowledge graph distillation}
\label{KGD-V}

As LKG captures language knowledge only, we further design vision knowledge graph distillation (KGD-V) that extracts a vision knowledge graph (VKG) and encapsulates it into the teacher detector to improve pseudo label generation. Specifically, VKG captures vision knowledge dynamically along the training process, which complement LKG by providing orthogonal and update-to-date vision information.

\textbf{Dynamic VKG Extraction. } We first initialize VKG with the CLIP text embedding and then employ the update-to-date object features to update it using manifold smoothing.
Specifically, we initialize VKG as a weighted undirected graph $G_{V} = ({V}_V, {U}_V)$, in which each node $\mathbf{v}_i\in {V}_V$ is initialized with the CLIP text embedding of category $\mathbf{c}_i$:
\begin{equation}\label{V-KGInit}
    \mathbf{v}_i = T\left( \mathbf{c}_i \right),
\end{equation}
and the graph edge $u_{ij}\in {U}_V$ is defined as the cosine similarity between nodes $\mathbf{v}_i$ and $\mathbf{v}_j$.
Given a batch of $\{\mathbf{x}_b^t\}_{b=1}^B \subseteq \mathcal{D}_t$ and the corresponding pseudo labels $\{\hat{\mathbf{y}}_b\}_{b=1}^B$ and CLIP features $\{ {F}_b\}_{b=1}^B$, the visual embedding centroid of category $\mathbf{c}_k$ can be obtained as the following:
\begin{equation}\label{VisualEmbeddingCentroidCal}
    \boldsymbol{\theta}_{i} = \frac{\mathop{\sum}_{b=1}^B \mathop{\sum}_{\mathbf{f}_j\in \mathbf{F}_b} \mathbf{f}_j \cdot \mathbb{I}(\hat{\mathbf{p}}_j(i)==\hat{\mathbf{p}}_j^{max})}{\mathop{\sum}_{b=1}^B \mathop{\sum}_{\mathbf{f}_j\in \mathbf{F}_b} \mathbb{I}(\hat{\mathbf{p}}_j(i)==\hat{\mathbf{p}}_j^{max})},
\end{equation}
where $\hat{\mathbf{p}}_j^{max}$ is the maximum element in probability vector $\hat{\mathbf{p}}_j$, $\mathbb{I}$ is the indicator function. 
And an affinity matrix $A$ can be calculated as $A_{ij}=exp(-r_{ij}^2/\sigma^2)$ and $A_{ii}=0$, where ${r}_{ij}=||\boldsymbol{\theta}_{i}-\boldsymbol{\theta}_{j}||_2$ 
and $\sigma^2=\text{Var}(r_{ij}^2)$.
In each iteration, the node of VKG is preliminarily updated as:
\begin{equation}\label{V-KGUpdateNode}
    \mathbf{v}_i \gets \lambda \mathbf{v}_i + (1-\lambda)\boldsymbol{\theta}_i.
\end{equation}
In order to incorporate the downstream visual graph knowledge into VKG, we perform additional steps to smooth the node of VKG, using the affinity matrix $A$ from the current batch as a guide:
\begin{equation}\label{V-KGUpdateStructure}
    \mathbf{v}_i = \sum\limits_{j}W_{ij}\mathbf{\mathbf{v}}_j,
\end{equation}
where $W=({I}-\alpha L)^{-1}$, $L=D^{-\frac{1}{2}} A D^{-\frac{1}{2}}$, $D_{ii}=\sum_jA_{ij}$, $\alpha$ is a scaling factor set as~\cite{velazquez2022closer}, and $I$ is the identity matrix.

\textbf{VKG Encapsulation} encapsulate the orthogonal and update-to-date vision knowledge in the extracted VKG into the teacher detector, which complements LKG and further improves pseudo label generation. 
With the extracted dynamic VKG $G_{V}$ and the object features $F$ in image $\mathbf{x}^t$, we encapsulate the extracted VKG into $\Phi_t$ in a similar way as the LKG Encapsulation as follows:
\begin{equation}\label{V-KGMeasurement}
    {\mathbf{p}}_{ji}^{v} = \hat{\mathbf{p}}_{ji} \cdot\frac{exp(cos\left\langle \mathbf{f}_j, \mathbf{v}_i\right\rangle)}{\sum_{i'}exp(cos\left\langle \mathbf{f}_j, \mathbf{v}_{i'}\right\rangle)},
\end{equation}
where $\hat{\mathbf{p}}_{ji}$ is the $i$-th element in vector $\hat{\mathbf{p}}_j$, denoting the predicted probability of category $\mathbf{c}_i$.
The first term in Eq.~\ref{V-KGMeasurement} is the prediction probability from the teacher model while the second term in Eq.~\ref{V-KGMeasurement} is the prediction probability from VKG.
${\mathbf{p}}_{ji}^{v}$ is the prediction probability calibrated by VKG. 

In this way, KGD-V extracts and encapsulates the VKG from CLIP into the teacher detector, further refining the detection pseudo labels of visual objects by conditioning on related visual nodes in VKG.

\subsection{Overall objective}
Finally, with the pseudo labels ${\mathbf{p}}_{j}^{l}$ and ${\mathbf{p}}_{j}^{v}$ generated from KGD-L and KGD-V respectively, the unsupervised training loss of KGD can be formulated as the following:
\begin{equation}\label{OverallObjective}
    \mathcal{L}_{KGD}= \sum\limits_{\mathbf{x}^t \in \mathcal{D}_t}\left({\mathcal{L}\left(\Phi_s(\mathbf{x}^t), \tilde{\mathbf{y}}\right)+\mathcal{L}_{LKG}(\mathbf{x}^t)}\right),
\end{equation}
where $\tilde{\mathbf{y}}=\{(\tilde{\mathbf{p}}_{j},\hat{\mathbf{t}}_j)\}_{j=1}^M$,  and $\tilde{\mathbf{p}}_j={N}({{\mathbf{p}}_{j}^{l}+{\mathbf{p}}_{j}^{v}})$. 
${N}(\cdot)$ normalizes data to range $[0,1]$.
The pseudo code of the proposed KGD is provided in appendix.

\section{Experiments}
This section presents experimental results.
Sections \ref{exp_dataset} and \ref{exp_impl} describe the dataset and implementation details. 
Section \ref{exp_result} presents the experiments across various downstream domain datasets.
Section \ref{exp_ablation} and Section \ref{exp_discussion} provide ablation studies and discuss different features of KGD.

\subsection{Datasets}
\label{exp_dataset}

We perform experiments on 11 object detection datasets that span different downstream domains including the object detection for autonomous driving~\cite{cordts2016cityscapes, neuhold2017mapillary}, autonomous driving under different weather and time-of-day conditions~\cite{yu2018bdd100k}, intelligent surveillance~\cite{8387876, yongqiang2021baai, 9573394}, 
common objects~\cite{everingham2015pascal, shao2019objects365}, 
and artistic illustration~\cite{8578623}.
More dataset details can be found in the Appendix.

\begin{table}[tb]
\caption{
Benchmarking over autonomous driving datasets under various weather and time conditions. 
\dag\; signifies that the methods employ WordNet to retrieve category definitions given category names, and CLIP to predict classification pseudo labels for objects. 
We adopt AP50 in evaluations.
The results of all methods are acquired with the same baseline~\cite{zhou2022detecting} as shown in the first row.
}
\renewcommand{\arraystretch}{0.5}
\centering
\resizebox{\linewidth}{!}{%
\begin{tabular}{l|c|c|ccccc|ccc}
\toprule
\multirow{2}{*}{Method}& \multirow{2}{*}{Cityscapes~\cite{cordts2016cityscapes}} &\multirow{2}{*}{Vistas~\cite{neuhold2017mapillary}} &\multicolumn{5}{c|}{BDD100K-weather~\cite{yu2018bdd100k}} &\multicolumn{3}{c}{BDD100K-time-of-day~\cite{yu2018bdd100k}} \\ 
\cmidrule{4-11} &&& rainy &snowy &overcast &cloudy &foggy &daytime &dawn\&dusk &night \\ 
\midrule
Detic~\cite{zhou2022detecting} (Baseline) &46.5&35.0&34.3 &33.5 &39.1 &42.0 &28.4     &39.2 &35.3 &28.5  \\
\midrule
MT~\cite{tarvainen2017mean}                  &49.1&35.7&34.3&34.2&39.9&41.7&28.9         &40.0&36.3& 28.5\\
MT~\cite{tarvainen2017mean}\dag              &50.0&36.6&35.0&35.3&40.9&43.0&29.8         &42.1&38.4&29.1 \\
SHOT~\cite{liang2020we}                &49.9&36.5&34.9&34.5&40.2&42.0&34.7         &40.5&36.1&26.7\\
SHOT~\cite{liang2020we}\dag            &50.8&37.4&36.1&35.7&41.8&44.1&35.6         &42.4&38.1&28.0\\
SFOD~\cite{li2021free}                &49.3&35.6&32.5&33.0&40.5&43.3&33.8         &40.8&36.0&28.9\\
SFOD~\cite{li2021free}\dag            &50.3&36.6&33.6&33.8&42.8&45.6&34.7         &43.4&37.9&30.1\\
HCL~\cite{huang2021model}                 &49.5&36.0&34.7&34.5& 40.4&42.2&30.8        &40.6&36.7&28.2 \\
HCL~\cite{huang2021model}\dag             &50.7&37.0&35.6&35.7&42.2&44.3&31.9         &42.9&38.6&29.5\\
IRG-SFDA~\cite{vs2022instance}            &50.6&36.4&35.0&35.3&40.7&42.6&{36.4}       &40.8&36.4&27.8 \\
IRG-SFDA~\cite{vs2022instance}\dag        &51.7&37.5&35.9&36.4&42.6&44.8&{36.7}&43.0&38.3&28.9\\
\textbf{KGD (Ours)} &\textbf{53.6}&\textbf{40.3}&\textbf{37.3}&\textbf{37.1}&\textbf{44.6}&\textbf{48.2}&\textbf{38.0}&\textbf{46.6}&\textbf{41.0}&\textbf{31.2}\\
\bottomrule
\end{tabular}
}
\label{tab:autonomous_weather}
\vspace{-2.0em}
\end{table}

\subsection{Implementation details}
\label{exp_impl}

We adopt Detic~\cite{zhou2022detecting} as LVD, where CenterNet2~\cite{zhou2021probabilistic} with Swin-B~\cite{liu2021swin} is pre-trained on LVIS \cite{gupta2019lvis} for object localization and ImageNet-21K \cite{deng2009imagenet} for object classification.
During adaption, the updating rate of EMA detector is set as $0.9999$.
The pseudo labels generated by the teacher detector with confidence greater than the threshold $\tau=0.25$ are selected for adaptation. 
We use AdamW~\cite{loshchilov2017decoupled} optimizer with initial learning rate $5\times10^{-6}$  and weight decay $10^{-4}$, and adopt a cosine learning rate schedule without warm-up iterations. 
The batch size is 2 and the image's shorter side is set to 640 while maintaining the aspect ratio unchanged.

\subsection{Results}
\label{exp_result}
Tables \ref{tab:autonomous_weather}-\ref{tab:surve_comobj_art} show the benchmarking of our methods with state-of-the-art domain adaptive detection methods.
As there are few prior studies on LVD adaptation, we compare our proposed method with state-of-the-art source-free domain adaptation methods for benchmarking, including Mean Teacher (MT)~\cite{tarvainen2017mean}, SHOT~\cite{liang2020we}, SFOD~\cite{li2021free}, HCL~\cite{huang2021model}, and IRG-SFDA~\cite{vs2022instance}.
For fair comparison, we incorporate CLIP~\cite{radford2021learning} and WordNet~\cite{miller1995wordnet} into the compared methods (marked with \dag). Specifically, we employ WordNet~\cite{miller1995wordnet} to generate category definitions given category names, and CLIP~\cite{radford2021learning} to predict pseudo labels for object classification.

\textbf{Object detection for autonomous driving.}
As Table \ref{tab:autonomous_weather} shows, the proposed KGD outperforms the baseline substantially over the general autonomous driving datasets Cityscapes and Vistas (with an average improvement of $6.20$ in AP50). KGD also outperforms the state-of-the-art by $2.35$ on average, demonstrating the superiority of KGD in adapting pretrained LVDs toward autonomous driving scenarios with substantial inter-domain discrepancy.
In addition, Table \ref{tab:autonomous_weather} shows experiments on autonomous driving data under various weather and time conditions. We can observe that KGD still achieves superior detection performance even though the unlabeled target data experience large style variation and severe quality degradation.
Further, the experiments show that KGD still outperforms the state-of-the-art clearly when CLIP and WordNet are incorporated, validating that the performance gain largely comes from our novel knowledge graph distillation instead of merely using CLIP and WordNet.

\begin{table}[tb]
\caption{
Benchmarking over common objects datasets, intelligent surveillance datasets, and  artistic illustration datasets. 
\dag\; signifies that the methods employ WordNet to retrieved category definitions given category names, and CLIP to predict classification pseudo labels for objects. 
We adopt AP50 in evaluations.
The results of all methods are acquired with the same baseline~\cite{zhou2022detecting} as shown in the first row.
}
\renewcommand{\arraystretch}{0.5}
\centering
\resizebox{\linewidth}{!}{
\begin{tabular}{l|cc|ccc|ccc}
\toprule
\multirow{2}{*}{Method}&\multicolumn{2}{c|}{Common Objects} &\multicolumn{3}{c|}{Intelligent Surveillance} &\multicolumn{3}{c}{Artistic Illustration} \\

\cmidrule{2-9}&VOC~\cite{everingham2015pascal}&Objects365~\cite{shao2019objects365}& MIO-TCD\cite{8387876} & BAAI~\cite{yongqiang2021baai} & VisDrone~\cite{9573394} &Clipart1k~\cite{8578623}      &Watercolor2k~\cite{8578623}   &Comic2k~\cite{8578623} \\
\midrule
Detic~\cite{zhou2022detecting} (Baseline)  &83.9&29.4&20.6&20.6 &19.0&61.0           &58.9           & 51.2\\
\midrule
MT~\cite{tarvainen2017mean}          &85.6&31.0&20.0&23.4&18.9&62.7           &58.4           &49.8\\
MT~\cite{tarvainen2017mean}\dag      &86.2&31.4&20.9&23.9&20.4&63.4           &59.6           &51.1\\
SHOT~\cite{liang2020we}             &84.0&30.7&21.2&22.5&19.4&61.3           &58.3           &50.4\\
SHOT~\cite{liang2020we}\dag         &84.5&31.2&22.3&23.3&20.9&62.3           &59.8           &52.1\\
SFOD~\cite{li2021free}              &85.5&31.6 &19.8&22.8&18.8&63.4           &58.2           &50.1\\
SFOD~\cite{li2021free}\dag          &86.2&32.0&21.0&23.1&20.2&64.6           &59.3           &51.8\\
HCL~\cite{huang2021model}           &85.8&31.8&20.5&23.6&18.8&63.1           &58.3           &52.3\\
HCL~\cite{huang2021model}\dag       &86.5&32.3&21.1&24.1&19.6&64.7           &59.7           &53.7\\
IRG-SFDA~\cite{vs2022instance}      &86.0&32.0&20.7&22.8&18.8&63.3           &60.8           &50.4\\
IRG-SFDA~\cite{vs2022instance}\dag  &86.3&32.3&21.6&23.7&20.0&65.0           &61.5           &52.0\\
\textbf{KGD (Ours)}                 &\textbf{86.9}&\textbf{34.4}&\textbf{24.6}  &\textbf{24.3} &\textbf{23.7} &\textbf{69.1}  &\textbf{63.5}  &\textbf{55.6}\\
\bottomrule
\end{tabular}
}
\label{tab:surve_comobj_art}
\vspace{-1.0em}
\end{table}

\textbf{Object detection for intelligent surveillance.}
The detection results on intelligent surveillance datasets are presented in Table \ref{tab:surve_comobj_art}.
Notably, the proposed KGD surpasses all other methods by significant margins, which underscores the effectiveness of KGD in adapting the pretrained LVD towards the challenging surveillance scenarios with considerable variations in camera lenses and angles.
The performance improvements achieved by KGD in this context demonstrate its effectiveness in exploring the unlabeled surveillance datasets by retrieving the classification knowledge of CLIP.

\textbf{Object detection for common objects.}
We evaluate the effectiveness of our KGD on the common object detection task using Pascal VOC and Objects365. 
Table \ref{tab:surve_comobj_art} reports the detection results, showcasing significant improvements over the {baseline} and outperforming state-of-the-arts, thereby highlighting the superiority of KGD. 
Besides, we can observe that the performance improvements on the Pascal VOC dataset and Objects365 dataset are not as significant as those in autonomous driving. This discrepancy is attributed to the relatively smaller domain gap between common objects and the pretraining dataset of LVD.

\textbf{Object detection for artistic illustration.}
Table \ref{tab:surve_comobj_art} reports the detection results on artistic illustration datasets. 
The proposed KGD outperforms all other methods by substantial margins, which highlights the effectiveness of KGD in adapting the pretrained large-vocabulary object detector towards artistic images that exhibit distinct domain gaps with natural images.

\begin{table}[tb]
\caption{
{Ablation studies of KGD} with Language Knowledge Graph Distillation (KGD-L) and Vision Knowledge Graph Distillation (KGD-V). 
The experiments are conducted on the Cityscapes dataset.
}
\renewcommand{\arraystretch}{0.5}
\tiny
\centering
\resizebox{0.9\linewidth}{!}{
\resizebox{\linewidth}{!}{
\begin{tabular}{l|ccc}
\toprule
Method & Language Knowledge Graph Distillation & Vision Knowledge Graph Distillation & AP50\\
\midrule
Detic~\cite{zhou2022detecting} ({Baseline})&&&46.5\\
\midrule
&\checkmark &&52.8\\
&&\checkmark&52.7\\
\textbf{KGD (Ours)} &\checkmark&\checkmark&\textbf{53.6} \\\bottomrule
\end{tabular}
}
}
\label{tab:abla_study}
\vspace{-1.0em}
\end{table}

\begin{table}[tb]
\renewcommand{\arraystretch}{0.5}
\caption{Comparisons with existing CLIP knowledge distillation methods on LVD adaptation.
For a fair comparison, we incorporate them with Mean Teacher Method (the columns with ‘MT+’). 
The results of all methods are acquired with the same baseline~\cite{zhou2022detecting} as shown in the first column.
}
\centering
\resizebox{\linewidth}{!}{
\begin{tabular}{l|cccccc}
\toprule
Method&Detic~\cite{zhou2022detecting} ({Baseline})&MT~\cite{tarvainen2017mean}&MT~\cite{tarvainen2017mean}+VILD~\cite{gu2021open}&MT~\cite{tarvainen2017mean}+RegionKD~\cite{rasheed2022bridging}&MT~\cite{tarvainen2017mean}+OADP~\cite{wang2023object}&\textbf{KGD (Ours)}\\
\midrule
AP50&46.5&49.1&50.6&50.2&50.2&\textbf{53.6}\\
\bottomrule
\end{tabular}
}
\label{tab:discus_comp}
\vspace{-2.0em}
\end{table}

\subsection{Ablation studies}
\label{exp_ablation}
In Table \ref{tab:abla_study}, we conducted ablation studies to assess the individual contribution of our proposed KGD-L and KGD-V on the task of LVD adaptation.
The pretrained LVD (i.e., Detic~\cite{zhou2022detecting} without adaptation) does not perform well due to the significant variations between its pre-training data and the downstream data,
As a comparison, either KGD-L or KGD-V brings significant performance improvements (i.e., +6.3 of AP50 and +6.2 of AP50 over the baseline), demonstrating both language and vision knowledge graphs built from CLIP can clearly facilitate the unsupervised adaptation of large-vocabulary object detectors.
The combination of KGD-L and KGD-V performs the best clearly, showing that our KGD-L and KGD-V are complementary by providing orthogonal language and vision knowledge for regularizing the unsupervised adaptation of LVDs.

\subsection{Discussion}
\label{exp_discussion}

\textbf{Language knowledge graph (LKG) Extraction strategies.}
Our proposed KGD-L introduces the WordNet~\cite{miller1995wordnet} to uncover the implicitly encoded language knowledge in CLIP~\cite{radford2021learning} and accordingly enables to build a category-discriminative and domain-generalizable Language Knowledge Graph (LKG) as described in Section~\ref{KGD-L}.
We examine the superiority of the proposed LKG Extraction with WordNet Hierarchy by comparing it with "LKG Extraction with category names" and "LKG Extraction with WordNet~\cite{miller1995wordnet} Synset definitions",
the former builds LKG directly with the category names from downstream datasets while the latter directly builds LKG using WordNet Synset definitions that are retrieved from the WordNet database with category names from downstream datasets.
As Table \ref{tab:strate_study_LKG} shows, both strategies achieve sub-optimal performance.
For “LKG Extraction with category names”, the category names are often ambiguous and less informative which degrades adaptation.
For "LKG Extraction with WordNet Synset definitions", the used  WordNet Synset definitions are more category-discriminative but they often have knowledge gaps with downstream data, limiting adaptation of the pretrained LVDs.
As a comparison, our proposed LKG Extraction with WordNet Hierarchy performs clearly better due to the guidance of Synset definitions as well as their hyponym sets that captures more comprehensive structural knowledge from the WordNet hierarchy which helps generate category-discriminative and domain-generalizable LKG and facilitates the adaption of LVDs towards downstream data effectively.

\begin{table}[tb]
\caption{
Study of different KGD-L strategies. 
The experiments are conducted on the Cityscapes dataset.
}
\renewcommand{\arraystretch}{0.5}
\vspace{-1.0em}
\centering
\resizebox{\linewidth}{!}{
\begin{tabular}{l|c|c|c|c}
\toprule
Method&LKG Extraction with category names&LKG Extraction with WordNet Synset definitions &LKG Extraction with WordNet Hierarchy&AP50\\
\midrule
Detic~\cite{zhou2022detecting} (Source only) &&&&46.5\\
\midrule
&\checkmark &&&51.9\\
\textbf{KGD-L only}&&\checkmark&&52.0\\
&&&\checkmark&\textbf{52.8} \\\bottomrule
\end{tabular}
}
\label{tab:strate_study_LKG}
\vspace{-2.0em}
\end{table}

\textbf{Language knowledge graph (LKG) Encapsulation strategies.}
Our proposed KGD-L encapsulates the comprehensive knowledge in the extracted LKG into the teacher detector to facilitate detection pseudo label generation as described in Section~\ref{KGD-L}.
We examine the superiority of the proposed LKG Encapsulation by comparing it with "LKG Encapsulation by Feature Distance", which directly calculate and normalize the feature distance between object proposal feature and LKG nodes, and calibrates the original prediction probability from the teacher model using the normalized feature distance.  
As Table \ref{tab:strate_study_LKG_en} shows, "LKG Encapsulation by Feature Distance" does not perform well in model adaptation, largely because it cannot effectively aggregate and capture semantic relationships and associations between different nodes in our extracted LKG. 
As a comparison, our proposed LKG Encapsulation shows clear improvements as the language information is adaptively aggregated along the training process stabilizes and improves the model adaptation, validating the performance gain largely comes from our novel LKG Encapsulation designs instead of merely using WordNet~\cite{miller1995wordnet} embedding.

\begin{table}[tb]
\caption{
Study of different KGD-L strategies. 
The experiments are conducted on the Cityscapes dataset.
}
\renewcommand{\arraystretch}{0.5}
\vspace{-1.0em}
\tiny
\centering
\begin{tabular}{l|c|c|c}
\toprule
Method&LKG Encapsulation by Feature Distance&LKG Encapsulation&AP50\\
\midrule
Detic~\cite{zhou2022detecting} (Source only) &&&46.5\\
\midrule
&\checkmark &&49.6\\
\textbf{KGD-L only}&&\checkmark&52.8\\
\bottomrule
\end{tabular}
\label{tab:strate_study_LKG_en}
\vspace{-1.0em}
\end{table}

\textbf{Vision knowledge graph distillation (KGD-V) strategies.}
Our proposed KGD-V captures the Dynamic vision knowledge graph (VKG) along the training as described in Section \ref{KGD-V}, which complements LKG by providing orthogonal and update-to-date vision information.
We examine the proposed Dynamic VKG Extraction by comparing it with "Static VKG Extraction" and "Dynamic VKG Extraction without Smooth".
The former builds a static VKG with CLIP features of image crops of objects that are predicted by the pretrained LVD before adaptation and it remains unchanged during the LVD adaptation process, 
while the latter updates the VKG with Eq.~(\ref{V-KGUpdateNode}) but without smooth (Eq.~\ref{V-KGUpdateStructure}).
As Table \ref{tab:strate_study_VKG} shows, "Static VKG Extraction" does not perform well in model adaptation, largely because the extracted static VKG is biased towards the pretraining datasets of the LVD and impedes domain-specific adaptation.
For "Dynamic VKG Extraction without Smooth", the nodes in VKG are updated with unlabeled downstream data in Eq.~(\ref{V-KGUpdateNode}), but the downstream visual graph knowledge is not effectively incorporated into VKG, which limits the adaptation of the pretrained LVD.
As a comparison, our proposed Dynamic VKG Extraction shows clear improvements as the update-to-date vision information extracted along the training process dynamically stabilizes and improves the model adaptation.

\begin{table}[tb]
\caption{
Studies of different KGD-V strategies. 
The experiments are conducted on the Cityscapes dataset.
}
\renewcommand{\arraystretch}{0.5}
\vspace{-1.0em}
\centering
\resizebox{\linewidth}{!}{
\begin{tabular}{l|c|c|c|c}
\toprule
Method&Static VKG Extraction&Dynamic VKG Extraction without Smooth&Dynamic VKG Extraction&AP50\\
\midrule
Detic~\cite{zhou2022detecting} (Source only) &&&&46.5\\
\midrule
&\checkmark &&&51.9 \\
&&\checkmark &&52.2 \\
\textbf{KGD-V only} &&&\checkmark&\textbf{52.7} \\\bottomrule
\end{tabular}
}
\label{tab:strate_study_VKG}
\vspace{-1.0em}
\end{table}

\textbf{Comparisons with existing CLIP knowledge distillation methods for detection.}
We compared our KGD with existing CLIP knowledge distillation methods designed for detection tasks.
Most existing methods achieve CLIP knowledge distillation by mimicking its feature space, such as VILD~\cite{gu2021open}, RegionKD~\cite{rasheed2022bridging}, and OADP~\cite{wang2023object}. 
Table \ref{tab:discus_comp} reports the experimental results over the Cityscapes dataset, which shows existing CLIP knowledge distillation methods do not perform well in adapting LVDs to downstream tasks. 
The main reason is that they merely align the feature space between LVDs and CLIP without considering the inherent semantic relationships between different object categories.
KGD also performs knowledge distillation but works for LVDs adaption effectively, largely because it works by extracting and encapsulating knowledge CLIP knowledge graphs which enables accurate object classification by leveraging relevant nodes in the knowledge graphs.

\begin{figure}[tb]
\begin{tabular}{p{2.9cm}p{2.9cm}p{2.9cm}p{2.9cm}}
\raisebox{-0.5\height}{\includegraphics[width=1.1\linewidth,height=0.6\linewidth]{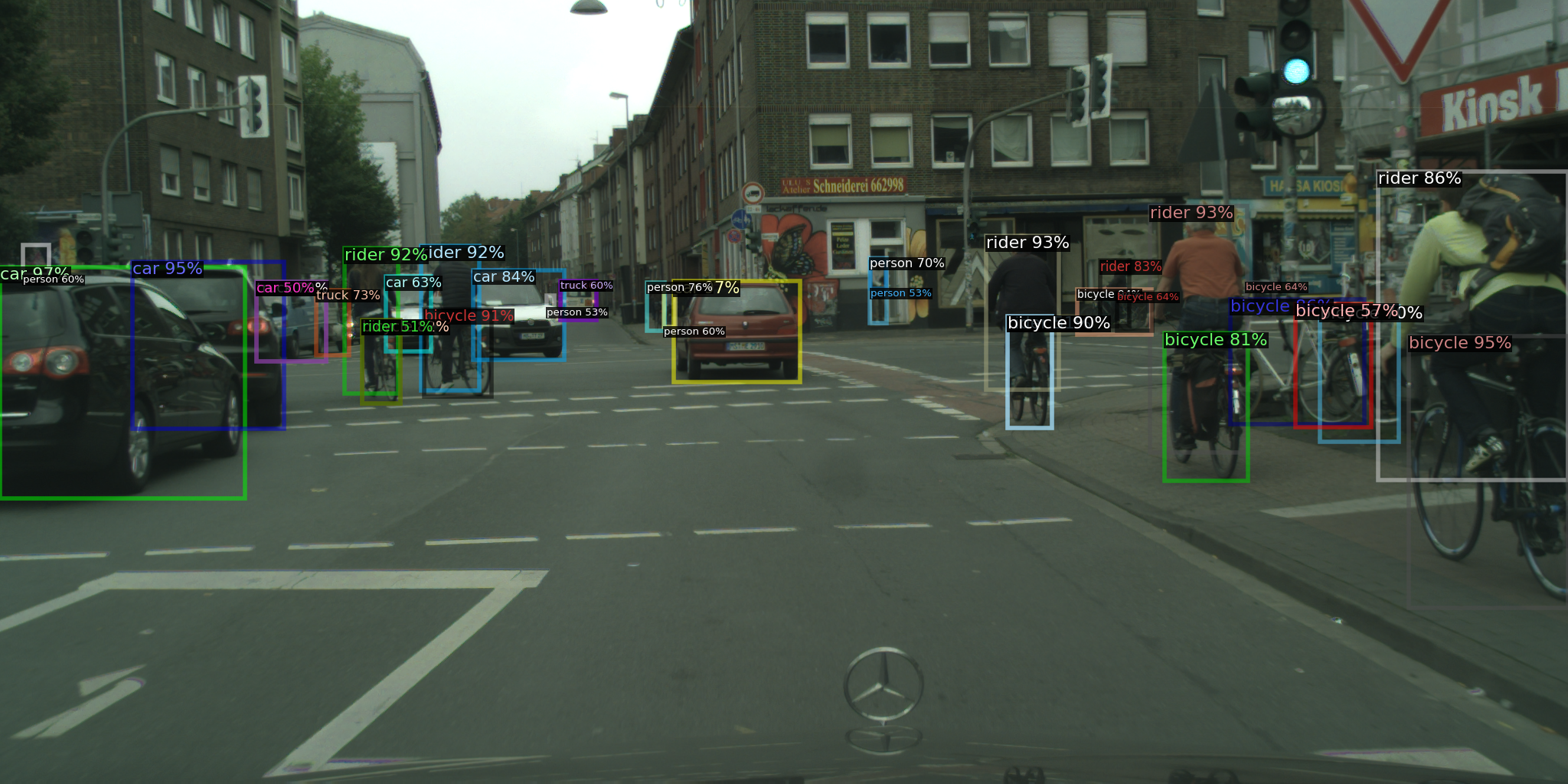}}
& \raisebox{-0.5\height}{\includegraphics[width=1.1\linewidth,height=0.6\linewidth]{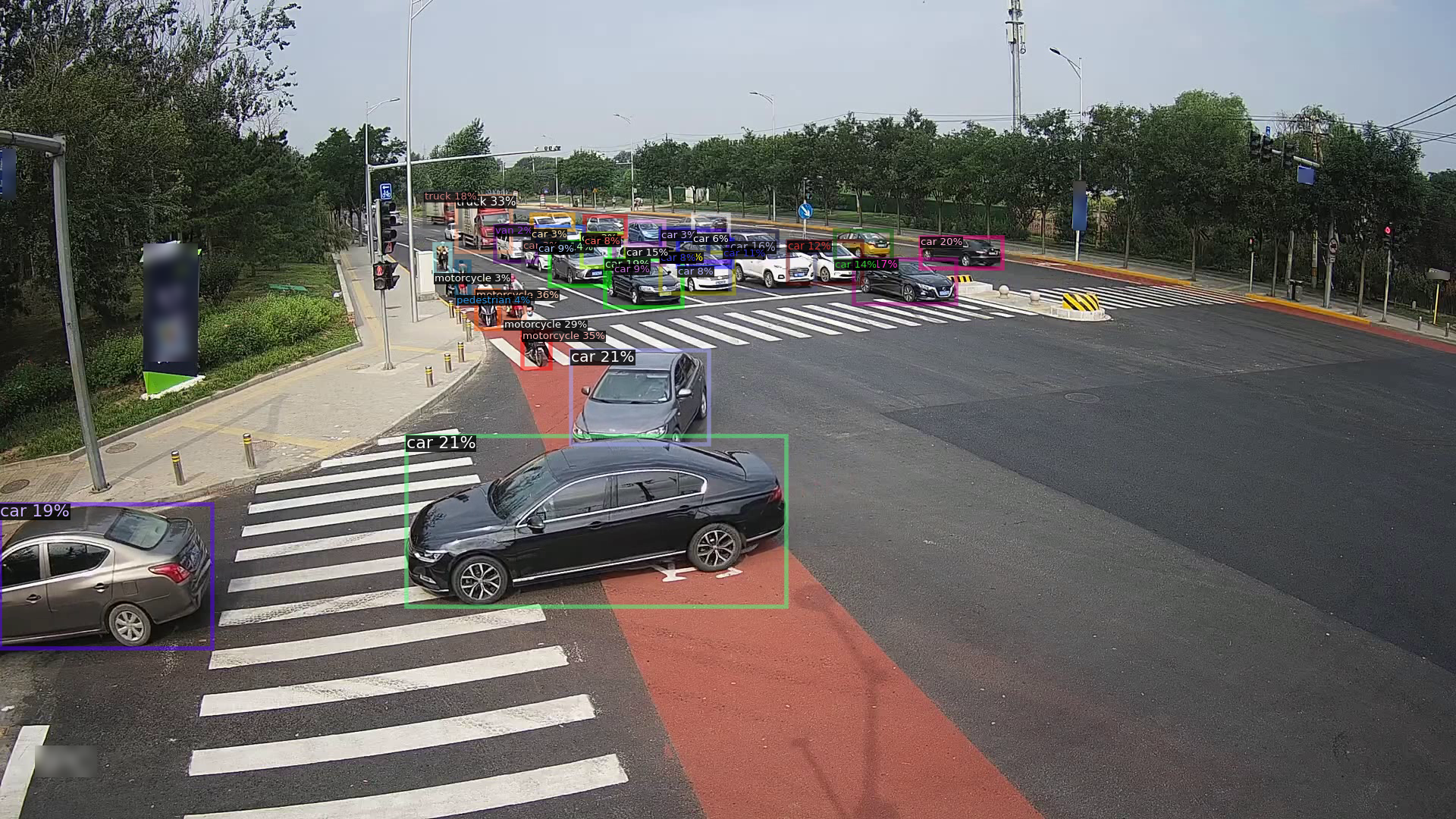}}
&\raisebox{-0.5\height}{\includegraphics[width=1.1\linewidth,height=0.6\linewidth]{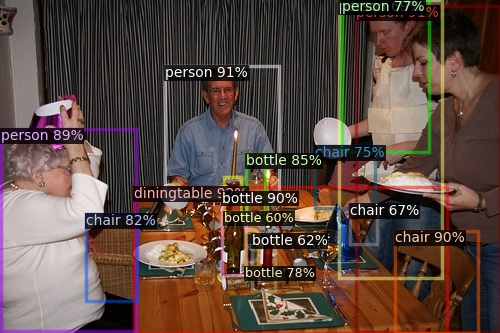}}
& \raisebox{-0.5\height}{\includegraphics[width=1.1\linewidth,height=0.6\linewidth]{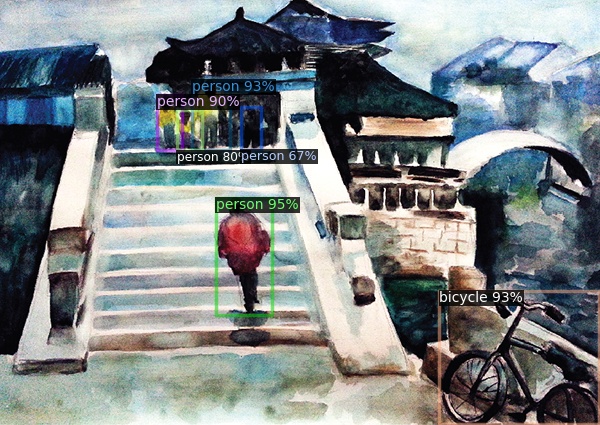}}
\\

 \raisebox{-0.5\height}{\centerline{\scriptsize{Autonomous driving}}}
& \raisebox{-0.5\height}{\centerline{\scriptsize{Intelligent surveillance}}}
& \raisebox{-0.5\height}{\centerline{\scriptsize{Common objects}}}
& \raisebox{-0.5\height}{\centerline{\scriptsize{Artistic illustration}}}
\\
\end{tabular}
\vspace{-1.0em}
\caption{
Qualitative comparisons over various datasets. Zoom in for details.
}
\label{fig:results}
\end{figure}

\textbf{Qualitative experimental results.}
We present qualitative results of KGD over diverse downstream domain detection datasets as shown in Fig.~\ref{fig:results}.
The qualitative results illustrate the effectiveness of KGD in producing accurate detection results across various domains, thereby qualitatively demonstrating its capability to adapt LVDs to unlabelled downstream domains with significant discrepancy in data distribution and vocabulary.

\begin{table}[tb]
\vspace{-1.0em}
\renewcommand{\arraystretch}{0.5}
\tiny
\caption{
Parameter analysis of KGD for the pseudo label generation threshold $\tau$.}
\centering
\begin{tabular}{c|ccccc}
\toprule
$\tau$ &0.15&0.2&0.25&0.3&0.35\\
\midrule
AP50   &53.4&53.2&53.6&53.9&53.5\\
\bottomrule
\end{tabular}
\label{tab:para_study}
\vspace{-2.0em}
\end{table}

\textbf{Parameter studies.} 
In the pseudo label generation in KGD, the reliable pseudo labels are acquired with a pre-defined confidence threshold $\tau$. 
We studied $\tau$ by changing it from $0.15$ to $0.35$ with a step of $0.05$. 
Table \ref{tab:para_study} reports the experiments over the Cityscapes dataset. It shows that $\tau$ does not affect KGD clearly, demonstrating the proposed KGD is tolerant to hyper-parameters.

\section{Conclusion}
This paper presents KGD, a novel knowledge distillation technique that exploits the implicit KG of CLIP to adapt large-vocabulary object detectors for handling various unlabelled downstream data. 
KGD consists of two consecutive stages including KG extraction and KG encapsulation which extract and encapsulate visual and textual KGs simultaneously, thereby providing complementary vision and language knowledge to facilitate unsupervised adaptation of large-vocabulary object detectors towards various downstream detection tasks. Extensive experiments on multiple widely-adopted detection datasets demonstrate that KGD consistently outperforms state-of-the-art techniques by clear margins.

%
%
\bibliographystyle{splncs04}
\bibliography{main}

\begin{thebibliography}{10}
\providecommand{\url}[1]{\texttt{#1}}
\providecommand{\urlprefix}{URL }
\providecommand{\doi}[1]{https://doi.org/#1}

\bibitem{alberts2020visualsem}
Alberts, H., Huang, T., Deshpande, Y., Liu, Y., Cho, K., Vania, C., Calixto, I.: Visualsem: a high-quality knowledge graph for vision and language. arXiv preprint arXiv:2008.09150  (2020)

\bibitem{chen2023pipa}
Chen, M., Zheng, Z., Yang, Y., Chua, T.S.: Pipa: Pixel-and patch-wise self-supervised learning for domain adaptative semantic segmentation. In: Proceedings of the 31st ACM International Conference on Multimedia. pp. 1905--1914 (2023)

\bibitem{chen2020self}
Chen, Y., Wei, C., Kumar, A., Ma, T.: Self-training avoids using spurious features under domain shift. Advances in Neural Information Processing Systems  \textbf{33},  21061--21071 (2020)

\bibitem{cordts2016cityscapes}
Cordts, M., Omran, M., Ramos, S., Rehfeld, T., Enzweiler, M., Benenson, R., Franke, U., Roth, S., Schiele, B.: The cityscapes dataset for semantic urban scene understanding. In: Proceedings of the IEEE conference on computer vision and pattern recognition. pp. 3213--3223 (2016)

\bibitem{dave2021evaluating}
Dave, A., Doll{\'a}r, P., Ramanan, D., Kirillov, A., Girshick, R.: Evaluating large-vocabulary object detectors: The devil is in the details. arXiv preprint arXiv:2102.01066  (2021)

\bibitem{deng2021gakg}
Deng, C., Jia, Y., Xu, H., Zhang, C., Tang, J., Fu, L., Zhang, W., Zhang, H., Wang, X., Zhou, C.: Gakg: A multimodal geoscience academic knowledge graph. In: Proceedings of the 30th ACM International Conference on Information \& Knowledge Management. pp. 4445--4454 (2021)

\bibitem{deng2009imagenet}
Deng, J., Dong, W., Socher, R., Li, L.J., Li, K., Fei-Fei, L.: Imagenet: A large-scale hierarchical image database. In: 2009 IEEE conference on computer vision and pattern recognition. pp. 248--255. Ieee (2009)

\bibitem{doveh2022teaching}
Doveh, S., Arbelle, A., Harary, S., Panda, R., Herzig, R., Schwartz, E., Kim, D., Giryes, R., Feris, R., Ullman, S., Karlinsky, L.: Teaching structured vision\&language concepts to vision\&language models (2022)

\bibitem{everingham2015pascal}
Everingham, M., Eslami, S.A., Van~Gool, L., Williams, C.K., Winn, J., Zisserman, A.: The pascal visual object classes challenge: A retrospective. International journal of computer vision  \textbf{111}(1),  98--136 (2015)

\bibitem{feng2021exploring}
Feng, C., Zhong, Y., Huang, W.: Exploring classification equilibrium in long-tailed object detection. In: Proceedings of the IEEE/CVF International conference on computer vision. pp. 3417--3426 (2021)

\bibitem{fürst2022cloob}
Fürst, A., Rumetshofer, E., Lehner, J., Tran, V., Tang, F., Ramsauer, H., Kreil, D., Kopp, M., Klambauer, G., Bitto-Nemling, A., Hochreiter, S.: Cloob: Modern hopfield networks with infoloob outperform clip (2022)

\bibitem{gao2022pyramidclip}
Gao, Y., Liu, J., Xu, Z., Zhang, J., Li, K., Ji, R., Shen, C.: Pyramidclip: Hierarchical feature alignment for vision-language model pretraining (2022)

\bibitem{gong2019dlow}
Gong, R., Li, W., Chen, Y., Gool, L.V.: Dlow: Domain flow for adaptation and generalization. In: Proceedings of the IEEE/CVF conference on computer vision and pattern recognition. pp. 2477--2486 (2019)

\bibitem{gu2021open}
Gu, X., Lin, T.Y., Kuo, W., Cui, Y.: Open-vocabulary object detection via vision and language knowledge distillation. arXiv preprint arXiv:2104.13921  (2021)

\bibitem{gupta2019lvis}
Gupta, A., Dollar, P., Girshick, R.: Lvis: A dataset for large vocabulary instance segmentation. In: Proceedings of the IEEE/CVF conference on computer vision and pattern recognition. pp. 5356--5364 (2019)

\bibitem{huang2021model}
Huang, J., Guan, D., Xiao, A., Lu, S.: Model adaptation: Historical contrastive learning for unsupervised domain adaptation without source data. Advances in Neural Information Processing Systems  \textbf{34} (2021)

\bibitem{8578623}
Inoue, N., Furuta, R., Yamasaki, T., Aizawa, K.: Cross-domain weakly-supervised object detection through progressive domain adaptation. In: 2018 IEEE/CVF Conference on Computer Vision and Pattern Recognition. pp. 5001--5009 (2018). \doi{10.1109/CVPR.2018.00525}

\bibitem{jia2021scaling}
Jia, C., Yang, Y., Xia, Y., Chen, Y.T., Parekh, Z., Pham, H., Le, Q.V., Sung, Y., Li, Z., Duerig, T.: Scaling up visual and vision-language representation learning with noisy text supervision (2021)

\bibitem{karim2023c}
Karim, N., Mithun, N.C., Rajvanshi, A., Chiu, H.p., Samarasekera, S., Rahnavard, N.: C-sfda: A curriculum learning aided self-training framework for efficient source free domain adaptation. In: Proceedings of the IEEE/CVF Conference on Computer Vision and Pattern Recognition. pp. 24120--24131 (2023)

\bibitem{kim2019self}
Kim, S., Choi, J., Kim, T., Kim, C.: Self-training and adversarial background regularization for unsupervised domain adaptive one-stage object detection. In: Proceedings of the IEEE International Conference on Computer Vision. pp. 6092--6101 (2019)

\bibitem{kim2021vilt}
Kim, W., Son, B., Kim, I.: Vilt: Vision-and-language transformer without convolution or region supervision. In: International Conference on Machine Learning. pp. 5583--5594. PMLR (2021)

\bibitem{lee2013pseudo}
Lee, D.H.: Pseudo-label: The simple and efficient semi-supervised learning method for deep neural networks. In: Workshop on Challenges in Representation Learning, ICML. vol.~3, p.~2 (2013)

\bibitem{li2022blip}
Li, J., Li, D., Xiong, C., Hoi, S.: Blip: Bootstrapping language-image pre-training for unified vision-language understanding and generation. In: International Conference on Machine Learning. pp. 12888--12900. PMLR (2022)

\bibitem{li2022grounded}
Li, L.H., Zhang, P., Zhang, H., Yang, J., Li, C., Zhong, Y., Wang, L., Yuan, L., Zhang, L., Hwang, J.N., et~al.: Grounded language-image pre-training. In: Proceedings of the IEEE/CVF Conference on Computer Vision and Pattern Recognition. pp. 10965--10975 (2022)

\bibitem{li2021free}
Li, X., Chen, W., Xie, D., Yang, S., Yuan, P., Pu, S., Zhuang, Y.: A free lunch for unsupervised domain adaptive object detection without source data. In: Proceedings of the AAAI Conference on Artificial Intelligence. vol.~35, pp. 8474--8481 (2021)

\bibitem{li2020overcoming}
Li, Y., Wang, T., Kang, B., Tang, S., Wang, C., Li, J., Feng, J.: Overcoming classifier imbalance for long-tail object detection with balanced group softmax. In: Proceedings of the IEEE/CVF conference on computer vision and pattern recognition. pp. 10991--11000 (2020)

\bibitem{liang2020we}
Liang, J., Hu, D., Feng, J.: Do we really need to access the source data? source hypothesis transfer for unsupervised domain adaptation. In: International Conference on Machine Learning. pp. 6028--6039. PMLR (2020)

\bibitem{liu2021cycle}
Liu, H., Wang, J., Long, M.: Cycle self-training for domain adaptation. Advances in Neural Information Processing Systems  \textbf{34},  22968--22981 (2021)

\bibitem{liu2021swin}
Liu, Z., Lin, Y., Cao, Y., Hu, H., Wei, Y., Zhang, Z., Lin, S., Guo, B.: Swin transformer: Hierarchical vision transformer using shifted windows. In: Proceedings of the IEEE/CVF international conference on computer vision. pp. 10012--10022 (2021)

\bibitem{lo2023spatio}
Lo, S.Y., Oza, P., Chennupati, S., Galindo, A., Patel, V.M.: Spatio-temporal pixel-level contrastive learning-based source-free domain adaptation for video semantic segmentation. In: Proceedings of the IEEE/CVF Conference on Computer Vision and Pattern Recognition. pp. 10534--10543 (2023)

\bibitem{long2015learning}
Long, M., Cao, Y., Wang, J., Jordan, M.: Learning transferable features with deep adaptation networks. In: International Conference on Machine Learning. pp. 97--105 (2015)

\bibitem{loshchilov2017decoupled}
Loshchilov, I., Hutter, F.: Decoupled weight decay regularization. arXiv preprint arXiv:1711.05101  (2017)

\bibitem{luo2021category}
Luo, Y., Liu, P., Zheng, L., Guan, T., Yu, J., Yang, Y.: Category-level adversarial adaptation for semantic segmentation using purified features. IEEE Transactions on Pattern Analysis and Machine Intelligence  \textbf{44}(8),  3940--3956 (2021)

\bibitem{8387876}
Luo, Z., Branchaud-Charron, F., Lemaire, C., Konrad, J., Li, S., Mishra, A., Achkar, A., Eichel, J., Jodoin, P.M.: Mio-tcd: A new benchmark dataset for vehicle classification and localization. IEEE Transactions on Image Processing  \textbf{27}(10),  5129--5141 (2018). \doi{10.1109/TIP.2018.2848705}

\bibitem{marino2021krisp}
Marino, K., Chen, X., Parikh, D., Gupta, A., Rohrbach, M.: Krisp: Integrating implicit and symbolic knowledge for open-domain knowledge-based vqa. In: Proceedings of the IEEE/CVF Conference on Computer Vision and Pattern Recognition. pp. 14111--14121 (2021)

\bibitem{mei2020instance}
Mei, K., Zhu, C., Zou, J., Zhang, S.: Instance adaptive self-training for unsupervised domain adaptation. In: Computer Vision--ECCV 2020: 16th European Conference, Glasgow, UK, August 23--28, 2020, Proceedings, Part XXVI 16. pp. 415--430. Springer (2020)

\bibitem{miller1995wordnet}
Miller, G.A.: Wordnet: a lexical database for english. Communications of the ACM  \textbf{38}(11),  39--41 (1995)

\bibitem{munir2021ssal}
Munir, M.A., Khan, M.H., Sarfraz, M., Ali, M.: Ssal: Synergizing between self-training and adversarial learning for domain adaptive object detection. Advances in Neural Information Processing Systems  \textbf{34},  22770--22782 (2021)

\bibitem{neuhold2017mapillary}
Neuhold, G., Ollmann, T., Rota~Bulo, S., Kontschieder, P.: The mapillary vistas dataset for semantic understanding of street scenes. In: Proceedings of the IEEE international conference on computer vision. pp. 4990--4999 (2017)

\bibitem{oksuz2020imbalance}
Oksuz, K., Cam, B.C., Kalkan, S., Akbas, E.: Imbalance problems in object detection: A review. IEEE transactions on pattern analysis and machine intelligence  \textbf{43}(10),  3388--3415 (2020)

\bibitem{pei2022hierarchical}
Pei, G., Shen, F., Yao, Y., Xie, G.S., Tang, Z., Tang, J.: Hierarchical feature alignment network for unsupervised video object segmentation (2022)

\bibitem{peng2023knowledge}
Peng, C., Xia, F., Naseriparsa, M., Osborne, F.: Knowledge graphs: Opportunities and challenges  (2023)

\bibitem{qu2022bmd}
Qu, S., Chen, G., Zhang, J., Li, Z., He, W., Tao, D.: Bmd: A general class-balanced multicentric dynamic prototype strategy for source-free domain adaptation. In: European Conference on Computer Vision. pp. 165--182. Springer (2022)

\bibitem{radford2021learning}
Radford, A., Kim, J.W., Hallacy, C., Ramesh, A., Goh, G., Agarwal, S., Sastry, G., Askell, A., Mishkin, P., Clark, J., et~al.: Learning transferable visual models from natural language supervision. In: International Conference on Machine Learning. pp. 8748--8763. PMLR (2021)

\bibitem{rasheed2022bridging}
Rasheed, H., Maaz, M., Khattak, M.U., Khan, S., Khan, F.S.: Bridging the gap between object and image-level representations for open-vocabulary detection. arXiv preprint arXiv:2207.03482  (2022)

\bibitem{redmon2017yolo9000}
Redmon, J., Farhadi, A.: Yolo9000: better, faster, stronger. In: Proceedings of the IEEE conference on computer vision and pattern recognition. pp. 7263--7271 (2017)

\bibitem{shao2019objects365}
Shao, S., Li, Z., Zhang, T., Peng, C., Yu, G., Zhang, X., Li, J., Sun, J.: Objects365: A large-scale, high-quality dataset for object detection. In: Proceedings of the IEEE/CVF international conference on computer vision. pp. 8430--8439 (2019)

\bibitem{sun2023domain}
Sun, T., Lu, C., Ling, H.: Domain adaptation with adversarial training on penultimate activations. In: Proceedings of the AAAI Conference on Artificial Intelligence. vol.~37, pp. 9935--9943 (2023)

\bibitem{tan2021equalization}
Tan, J., Lu, X., Zhang, G., Yin, C., Li, Q.: Equalization loss v2: A new gradient balance approach for long-tailed object detection. In: Proceedings of the IEEE/CVF conference on computer vision and pattern recognition. pp. 1685--1694 (2021)

\bibitem{tan2020equalization}
Tan, J., Wang, C., Li, B., Li, Q., Ouyang, W., Yin, C., Yan, J.: Equalization loss for long-tailed object recognition. In: Proceedings of the IEEE/CVF conference on computer vision and pattern recognition. pp. 11662--11671 (2020)

\bibitem{tarvainen2017mean}
Tarvainen, A., Valpola, H.: Mean teachers are better role models: Weight-averaged consistency targets improve semi-supervised deep learning results. In: Advances in neural information processing systems. pp. 1195--1204 (2017)

\bibitem{velazquez2022closer}
Velazquez, D., Rodr{\'\i}guez, P., Gonfaus, J.M., Roca, F.X., Gonz{\`a}lez, J.: A closer look at embedding propagation for manifold smoothing. The Journal of Machine Learning Research  \textbf{23}(1),  11447--11473 (2022)

\bibitem{vs2022instance}
VS, V., Oza, P., Patel, V.M.: Instance relation graph guided source-free domain adaptive object detection. arXiv preprint arXiv:2203.15793  (2022)

\bibitem{vu2019advent}
Vu, T.H., Jain, H., Bucher, M., Cord, M., P{\'e}rez, P.: Advent: Adversarial entropy minimization for domain adaptation in semantic segmentation. In: Proceedings of the IEEE Conference on Computer Vision and Pattern Recognition. pp. 2517--2526 (2019)

\bibitem{wang2021seesaw}
Wang, J., Zhang, W., Zang, Y., Cao, Y., Pang, J., Gong, T., Chen, K., Liu, Z., Loy, C.C., Lin, D.: Seesaw loss for long-tailed instance segmentation. In: Proceedings of the IEEE/CVF conference on computer vision and pattern recognition. pp. 9695--9704 (2021)

\bibitem{wang2023object}
Wang, L., Liu, Y., Du, P., Ding, Z., Liao, Y., Qi, Q., Chen, B., Liu, S.: Object-aware distillation pyramid for open-vocabulary object detection. arXiv preprint arXiv:2303.05892  (2023)

\bibitem{wang2021domain}
Wang, Q., Dai, D., Hoyer, L., Van~Gool, L., Fink, O.: Domain adaptive semantic segmentation with self-supervised depth estimation. In: Proceedings of the IEEE/CVF International Conference on Computer Vision. pp. 8515--8525 (2021)

\bibitem{wilcke2020end}
Wilcke, W., Bloem, P., de~Boer, V., van~t Veer, R., van Harmelen, F.: End-to-end entity classification on multimodal knowledge graphs. arXiv preprint arXiv:2003.12383  (2020)

\bibitem{wu2020forest}
Wu, J., Song, L., Wang, T., Zhang, Q., Yuan, J.: Forest r-cnn: Large-vocabulary long-tailed object detection and instance segmentation. In: Proceedings of the 28th ACM International Conference on Multimedia. pp. 1570--1578 (2020)

\bibitem{wu2020comprehensive}
Wu, Z., Pan, S., Chen, F., Long, G., Zhang, C., Philip, S.Y.: A comprehensive survey on graph neural networks. IEEE transactions on neural networks and learning systems  \textbf{32}(1),  4--24 (2020)

\bibitem{yang2019detecting}
Yang, H., Wu, H., Chen, H.: Detecting 11k classes: Large scale object detection without fine-grained bounding boxes. In: Proceedings of the IEEE/CVF International Conference on Computer Vision. pp. 9805--9813 (2019)

\bibitem{Yang_2019_ICCV}
Yang, H., Wu, H., Chen, H.: Detecting 11k classes: Large scale object detection without fine-grained bounding boxes. In: Proceedings of the IEEE/CVF International Conference on Computer Vision (ICCV) (October 2019)

\bibitem{yang2021st3d}
Yang, J., Shi, S., Wang, Z., Li, H., Qi, X.: St3d: Self-training for unsupervised domain adaptation on 3d object detection. In: Proceedings of the IEEE/CVF conference on computer vision and pattern recognition. pp. 10368--10378 (2021)

\bibitem{yao2022detclip}
Yao, L., Han, J., Wen, Y., Liang, X., Xu, D., Zhang, W., Li, Z., Xu, C., Xu, H.: Detclip: Dictionary-enriched visual-concept paralleled pre-training for open-world detection. arXiv preprint arXiv:2209.09407  (2022)

\bibitem{yao2021filip}
Yao, L., Huang, R., Hou, L., Lu, G., Niu, M., Xu, H., Liang, X., Li, Z., Jiang, X., Xu, C.: Filip: Fine-grained interactive language-image pre-training (2021)

\bibitem{yi2023source}
Yi, L., Xu, G., Xu, P., Li, J., Pu, R., Ling, C., McLeod, A.I., Wang, B.: When source-free domain adaptation meets learning with noisy labels. arXiv preprint arXiv:2301.13381  (2023)

\bibitem{yongqiang2021baai}
Yongqiang, D., Dengjiang, W., Gang, C., Bing, M., Xijia, G., Yajun, W., Jianchao, L., Yanming, F., Juanjuan, L.: Baai-vanjee roadside dataset: Towards the connected automated vehicle highway technologies in challenging environments of china. arXiv preprint arXiv:2105.14370  (2021)

\bibitem{you2019universal}
You, K., Long, M., Cao, Z., Wang, J., Jordan, M.I.: Universal domain adaptation. In: Proceedings of the IEEE/CVF conference on computer vision and pattern recognition. pp. 2720--2729 (2019)

\bibitem{yu2018bdd100k}
Yu, F., Xian, W., Chen, Y., Liu, F., Liao, M., Madhavan, V., Darrell, T.: Bdd100k: A diverse driving video database with scalable annotation tooling. arXiv preprint arXiv:1805.04687  \textbf{2}(5), ~6 (2018)

\bibitem{yue2021prototypical}
Yue, X., Zheng, Z., Zhang, S., Gao, Y., Darrell, T., Keutzer, K., Vincentelli, A.S.: Prototypical cross-domain self-supervised learning for few-shot unsupervised domain adaptation. In: Proceedings of the IEEE/CVF Conference on Computer Vision and Pattern Recognition. pp. 13834--13844 (2021)

\bibitem{zhang2023visionlanguage}
Zhang, J., Huang, J., Jin, S., Lu, S.: Vision-language models for vision tasks: A survey. arXiv preprint arXiv:2304.00685  (2023)

\bibitem{zhang2021prototypical}
Zhang, P., Zhang, B., Zhang, T., Chen, D., Wang, Y., Wen, F.: Prototypical pseudo label denoising and target structure learning for domain adaptive semantic segmentation. In: Proceedings of the IEEE/CVF conference on computer vision and pattern recognition. pp. 12414--12424 (2021)

\bibitem{zhang2018adaptive}
Zhang, Q., Fu, J., Liu, X., Huang, X.: Adaptive co-attention network for named entity recognition in tweets. In: Proceedings of the AAAI conference on artificial intelligence. vol.~32 (2018)

\bibitem{zhang2019category}
Zhang, Q., Zhang, J., Liu, W., Tao, D.: Category anchor-guided unsupervised domain adaptation for semantic segmentation. In: Advances in Neural Information Processing Systems. pp. 433--443 (2019)

\bibitem{zhang2021distribution}
Zhang, S., Li, Z., Yan, S., He, X., Sun, J.: Distribution alignment: A unified framework for long-tail visual recognition. In: Proceedings of the IEEE/CVF conference on computer vision and pattern recognition. pp. 2361--2370 (2021)

\bibitem{zhang2022divide}
Zhang, Z., Chen, W., Cheng, H., Li, Z., Li, S., Lin, L., Li, G.: Divide and contrast: Source-free domain adaptation via adaptive contrastive learning. Advances in Neural Information Processing Systems  \textbf{35},  5137--5149 (2022)

\bibitem{zhou2022detecting}
Zhou, X., Girdhar, R., Joulin, A., Kr{\"a}henb{\"u}hl, P., Misra, I.: Detecting twenty-thousand classes using image-level supervision. arXiv preprint arXiv:2201.02605  (2022)

\bibitem{zhou2021probabilistic}
Zhou, X., Koltun, V., Kr{\"a}henb{\"u}hl, P.: Probabilistic two-stage detection. arXiv preprint arXiv:2103.07461  (2021)

\bibitem{9573394}
Zhu, P., Wen, L., Du, D., Bian, X., Fan, H., Hu, Q., Ling, H.: Detection and tracking meet drones challenge. IEEE Transactions on Pattern Analysis and Machine Intelligence pp.~1--1 (2021). \doi{10.1109/TPAMI.2021.3119563}

\bibitem{zhu2022multi}
Zhu, X., Li, Z., Wang, X., Jiang, X., Sun, P., Wang, X., Xiao, Y., Yuan, N.J.: Multi-modal knowledge graph construction and application: A survey. arXiv preprint arXiv:2202.05786  (2022)

\bibitem{zou2021geometry}
Zou, L., Tang, H., Chen, K., Jia, K.: Geometry-aware self-training for unsupervised domain adaptation on object point clouds. In: Proceedings of the IEEE/CVF International Conference on Computer Vision. pp. 6403--6412 (2021)

\bibitem{zou2018unsupervised}
Zou, Y., Yu, Z., Vijaya~Kumar, B., Wang, J.: Unsupervised domain adaptation for semantic segmentation via class-balanced self-training. In: Proceedings of the European Conference on Computer Vision (ECCV). pp. 289--305 (2018)

\end{thebibliography}
\end{document}